\documentclass[10pt,conference,a4paper]{IEEEtran}

\usepackage{graphicx}
\usepackage{caption}
\usepackage{subcaption}
\usepackage{amsmath}
\usepackage{algorithmic}
\usepackage{algorithm}
\usepackage{lineno}
\usepackage{multirow}
\usepackage{tabularx}
\usepackage{booktabs}
\usepackage{url}
\usepackage{color}
\usepackage[figuresright]{rotating}
\usepackage[utf8]{inputenc}
\usepackage[T1]{fontenc}

\newcolumntype{Y}{>{\arraybackslash}X}

\begin{document}

\title{CSMOUTE: Combined Synthetic Oversampling and Undersampling Technique for Imbalanced Data Classification}

\author{\IEEEauthorblockN{Michał Koziarski}
\IEEEauthorblockA{Department of Electronics\\
AGH University of Science and Technology\\
Al. Mickiewicza 30, 30-059 Kraków, Poland\\
Email: michal.koziarski@agh.edu.pl}}

\maketitle

\begin{abstract}
In this paper we propose a novel data-level algorithm for handling data imbalance in the classification task, Synthetic Majority Undersampling Technique (SMUTE). SMUTE leverages the concept of interpolation of nearby instances, previously introduced in the oversampling setting in SMOTE. Furthermore, we combine both in the Combined Synthetic Oversampling and Undersampling Technique (CSMOUTE), which integrates SMOTE oversampling with SMUTE undersampling. The results of the conducted experimental study demonstrate the usefulness of both the SMUTE and the CSMOUTE algorithms, especially when combined with more complex classifiers, namely MLP and SVM, and when applied on datasets consisting of a large number of outliers. This leads us to a conclusion that the proposed approach shows promise for further extensions accommodating local data characteristics, a direction discussed in more detail in the paper.
\end{abstract}

\IEEEpeerreviewmaketitle

\section{Introduction}

Data imbalance remains one of the open challenges of the contemporary machine learning, affecting, to some extent, a majority of the real-world classification problems. It occurs whenever one of the considered classes, a so-called \textit{majority class}, consists of a higher number of observations than one of the other \textit{minority classes}. Data imbalance is especially prevalent in problem domains in which the data acquisition for the minority class poses a greater difficulty, and the majority class observations are more abundant. Examples of such domains include, but are not limited to, cancer malignancy grading \cite{koziarski2018convolutional}, fraud detection \cite{wei2013effective}, behavioral analysis \cite{azaria2014behavioral} and cheminformatics \cite{czarnecki2015compounds}. 

Data imbalance poses a challenge for traditional learning algorithms, which are ill-equipped for handling uneven class distributions and tend to display a bias toward the majority class accompanied by a reduced discriminatory capabilities on the minority classes. A large variety of methods reducing the negative impact of data imbalance on the classification performance can be found in the literature. They can be divided into two categories, based on the part of the classification pipeline that they modify. First of all, the \textit{data-level} algorithms, in which the training data is manipulated prior to the classification by either reducing the number of majority observations (\textit{undersampling}) or increasing the number of minority observations (\textit{oversampling}). Secondly, the \textit{algorithm-level} methods, which adjust the training procedure of the learning algorithms to better accommodate for the data imbalance. In this paper we focus on the former. Specifically, we propose a novel undersampling algorithm, Synthetic Majority Undersampling Technique (SMUTE), which leverages the concept of interpolation of nearby instances, previously introduced in the oversampling setting in SMOTE \cite{chawla2002smote}. Secondly, we propose a Combined Synthetic Oversampling and Undersampling Technique (CSMOUTE), which integrates SMOTE oversampling with SMUTE undersampling. The aim of this paper is to serve as a preliminary study of the potential usefulness of the proposed approach, with the final goal of extending it to utilize local data characteristics, a direction further discussed in the remainder of the paper. To this end, in the conducted experiments we not only compare the proposed method with the state-of-the-art approaches, but also analyse the factors influencing its performance, with a particular focus on the impact of dataset characteristics.

\section{Related Work}

Various different approaches to the imbalanced data undersampling can be distinguished in the literature. Perhaps the oldest techniques are heuristic cleaning strategies such as Tomek links \cite{tomek1976two}, Edited Nearest-Neighbor rule \cite{wilson1972asymptotic}, Condensed Nearest Neighbour editing (CNN) \cite{hart1968condensed}, and more recently Near Miss method (NM) \cite{mani2003knn}. They tend not to allow specifying a desired undersampling ratio, instead removing all of the instances meeting a predefined criterion. This can lead to an undesired behavior in the cases in which the imbalance level after undersampling still does not meet users' expectation. As a result, contemporary methods tend to allow an arbitrary level of balancing. This can be achieved by introducing a scoring function and removing the majority observations based on the order it introduces. For instance, Anand et al. \cite{anand2010approach} propose sorting the undersampled observations based on the weighted Euclidean distance from the positive samples. Smith et al. \cite{smith2014instance} advocate for using the instance hardness criterion, with the hardness estimated based on the certainty of the classifiers predictions. Another approach that allows specifying the desired level of undersampling are clustering-based approaches, which reduce the number of original observations by replacing them with a specified number of representative prototypes \cite{yen2009cluster,beckmann2015knn}. Finally, as has been originally demonstrated by Liu et al. \cite{liu2008exploratory}, undersampling algorithms are well-suited for forming classifier ensembles, an idea that was further extended in form of evolutionary undersampling \cite{galar2013eusboost} and boosting \cite{lu2017adaptive}.

Over- and undersampling strategies for handling data imbalance pose unique challenges during the algorithm design process, and can lead to a vastly different performance on any given dataset. Some research has been done on the factors affecting the relative performance of both of these approaches. First of all, some of the classification algorithms show clear preference towards either of the resampling strategies, with a notable example of decision trees, the overfitting of which was a motivation behind the SMOTE \cite{chawla2002smote}. Nevertheless, later study found the SMOTE itself to still be ineffective when combined with the C4.5 algorithm \cite{drummond2003c4}, for which applying undersampling led to a better performance. In another study \cite{van2009knowledge} authors focused on the impact of noise, with a conclusion that especially for high levels of noise simple random undersampling produced the best results. Finally, in another study \cite{garcia2012effectiveness} authors investigated the impact of the level of imbalance on the choice of the resampling strategy. Their results indicate that oversampling tends to perform better on severely imbalanced datasets, while for more modest levels of imbalance both over- and undersampling tend achieve similar results. In general, none of the approaches is clearly outperforming the other, and both can be useful in specific cases. As a result, a family of methods combining over- and undersampling emerged. One of the first of such approaches involved combining SMOTE with later cleaning of the complete dataset, using either Tomek links \cite{tomek1976two} or Edited Nearest-Neighbor rule \cite{wilson1972asymptotic}. More recently, methods such as SMOUTE \cite{songwattanasiri2010smoute}, which combines SMOTE oversampling with $k$-means based undersampling, as well as CORE \cite{bunkhumpornpat2015core}, a technique the goal of which is to strengthen the core of a minority class while simultaneously reducing the risk of incorrect classification of borderline instances, were proposed. In another study \cite{junsomboon2017combining} authors propose combining SMOTE with Neighborhood Cleaning Rule (NCL) \cite{laurikkala2001improving}. In general, based on the results present in the literature combining over- and undersampling tends to improve the performance of the underlying algorithms. However, to the best of our knowledge no dedicated analysis of the factors under which combined resampling outperforms over- and undersampling has been published.

One important factor that can influence the suitability of a method for a given dataset are the characteristics of the minority class observations it contains. In a study by Napierała and Stefanowski \cite{napierala2016types} authors proposed a method for categorization of different types of minority objects. Their approach used a 5-neighborhood to identify the nearest neighbors of a given observation, and afterwards assign to it a category based on the proportion of neighbors from the same class, as either safe, borderline, rare or outlier. In a recent study by Koziarski \cite{koziarski2020radial} this categorization was used to analyse the properties of the proposed Radial-Based Undersampling algorithm and identify the types of datasets on which it achieves the best results. Similar categorization was also used in the design of a number of over- and undersampling approaches, in which the resampling was focused on observations of a particular type. These include several extensions of SMOTE, such as Borderline-SMOTE \cite{han2005borderline}, focusing on the borderline instances, placed close to the decision border, and Safe-Level-SMOTE \cite{Bunkhumpornpat:2009} and LN-SMOTE \cite{Maciejewski:2011}, limiting the risk of placing synthetic instances inside the regions belonging to the majority class, as well as MUTE \cite{bunkhumpornpat2011mute}, extending the  concept of Safe-Level-SMOTE to the undersampling setting. However, these methods tend to be ad-hoc and in most cases their behavior on datasets of a specific type is not analysed. One important exception is a study conducted by Sáez et al. \cite{saez2016analyzing}, in which authors used the extracted knowledge about the imbalance distribution types to guide the oversampling process. 

\section{CSMOUTE Algorithm}

To mitigate the negative impact of data imbalance on the performance of classification algorithms we propose a novel data-level approach, Synthetic Majority Undersampling Technique (SMUTE), which leverages the concept of interpolation of nearby instances in the undersampling process. The idea of using interpolation between nearby instances was previously introduced in the oversampling setting as SMOTE \cite{chawla2002smote} and since then became a cornerstone of numerous data-level strategies for handling data imbalance \cite{fernandez2018smote}. SMOTE was introduced as a direct response to the shortcomings of random oversampling, which was shown by Chawla et al. to cause overfitting in selected classification algorithms, such as decision trees. Instead of simply duplicating the minority instances, which was the case in random oversampling, SMOTE instead advocates for generating synthetic observations via data interpolation. It is worth noting that this approach does not remain without its on disadvantages. For instance, SMOTE does not take into the account the positions of the majority class objects, and as a result can produce synthetic observations overlapping the majority class distribution. This was further discussed by Koziarski et al. \cite{koziarski2019radial}. Still, despite its flaws, SMOTE remains one of the most important approaches for handling data imbalance. 

In this paper we propose translating the procedure of generating synthetic instances via interpolation of nearby observations to the undersampling setting. Specifically, in the proposed algorithm we focus on the majority class observations, and at each iteration randomly select one of them and one of their $k$ nearest neighbors, with $k$ being a parameter of the algorithm. Afterwards, we synthesize a new instance, similar to the SMOTE, and add it to the collection of majority observations, but at the same time remove two of the original instances from the dataset. The pseudocode of the proposed approach was presented in Algorithm~\ref{algorithm:csmoute}. In contrast to the SMOTE, the proposed procedure is not motivated by the need to combat overfitting, which is not as pressing an issue in the case of undersampling, but instead the one to minimize the loss of information due to the instance removal. The idea is similar to various prototype selection approaches, in which we modify the original collection of majority observations by a smaller collection of representative instances, for instance by applying clustering. In fact, the proposed approach can be viewed as a special case of a local clustering, in which instead of using the whole set of majority observations we select only its subset, in this case an extreme collection consisting only of two nearby observations, and cluster them at the same time adding a random noise to the resulting synthetic observation. However, in contrast with the traditional clustering approaches, likely to operate on either a whole collection of majority observations, or its much larger subset, proposed approach differs in two significant ways. First of all, by using only two observations during the reduction step we reduce the capabilities of the algorithm to generate representative prototypes, an obvious drawback of the method. However, since the reduction step originates in the position of a particular existing observation, it allows us to guide the resampling procedure. Specifically, it enables us to select only the points of origin with particular characteristics, for instance only the borderline instances, or only the outliers, which is either not possible or not as easily achievable in the traditional clustering approaches. Similarly, it allows us to adjust the ratio of points of origin of a given type what will be used for reduction, or accommodate other local data characteristics. This is further facilitated by the second of the proposed approaches, the Combined Synthetic Oversampling and Undersampling Technique (CSMOUTE), which simply combines SMOTE oversampling with SMUTE undersampling at a given ratio of over- and undersampling. While we do not explicitly evaluate the idea of using local data characteristics in the process of CSMOUTE resampling, this is the main motivation behind using the proposed approach instead of other, clustering-based undersampling algorithms, and the goal of the evaluation conducted in the remainder of this paper is establishing whether CSMOUTE, on its own, constitutes a feasible enough method for further extensions.

\begin{algorithm}
\caption{CSMOUTE}
\label{algorithm:csmoute}
\begin{algorithmic}[1]
\STATE \textbf{Input:} collections of majority observations $\mathcal{X}_{maj}$ and minority observations $\mathcal{X}_{min}$
\STATE \textbf{Parameters:} number of nearest neighbors used during oversampling $k_{SMOTE}$ and undersampling $k_{SMUTE}$, $ratio$ of data balancing done using oversampling
\STATE \textbf{Output:} resampled collections of majority observations $\mathcal{X}_{maj}'$ and minority observations $\mathcal{X}_{min}'$
\STATE
\STATE \textbf{function} SMOTE($k$, $n$):
\STATE $\mathcal{X}_{min}' \gets \mathcal{X}_{min}$
\WHILE{$\vert \mathcal{X}_{min}' \vert - \vert \mathcal{X}_{min} \vert < n$}
\STATE $x_1 \gets $ randomly selected observation from $\mathcal{X}_{min}$
\STATE $x_2 \gets $ randomly selected observation from $k$ nearest neighbors of $x_1$ belonging to $\mathcal{X}_{min}$
\STATE $r \gets $ random real number in $[0, 1]$
\STATE $x' \gets x_1 + r \cdot (x_2 - x_1)$
\STATE append $x'$ to $\mathcal{X}_{min}'$
\ENDWHILE
\STATE \textbf{return} $\mathcal{X}_{min}'$
\STATE
\STATE \textbf{function} SMUTE($k$, $n$):
\STATE $\mathcal{X}_{maj}' \gets \mathcal{X}_{maj}$
\WHILE{$\vert \mathcal{X}_{maj} \vert - \vert \mathcal{X}_{maj}' \vert < n$}
\STATE $x_1 \gets $ randomly selected observation from $\mathcal{X}_{maj}'$
\STATE $x_2 \gets $ randomly selected observation from $k$ nearest neighbors of $x_1$ belonging to $\mathcal{X}_{maj}'$
\STATE $r \gets $ random real number in $[0, 1]$
\STATE $x' \gets x_1 + r \cdot (x_2 - x_1)$
\STATE delete $x_1$ and $x_2$ from $\mathcal{X}_{maj}'$
\STATE append $x'$ to $\mathcal{X}_{maj}'$
\ENDWHILE
\STATE \textbf{return} $\mathcal{X}_{maj}'$
\STATE
\STATE \textbf{function} CSMOUTE($k_{SMOTE}$, $k_{SMUTE}$, $ratio$):
\STATE $n \gets \vert \mathcal{X}_{maj} \vert - \vert \mathcal{X}_{min} \vert$
\STATE $n_{SMOTE} \gets \textrm{round}(n \cdot ratio)$
\STATE $n_{SMUTE} \gets n - n_{SMOTE}$
\STATE $\mathcal{X}_{min}' \gets \textrm{SMOTE}(k_{SMOTE}, n_{SMOTE})$
\STATE $\mathcal{X}_{maj}' \gets \textrm{SMUTE}(k_{SMUTE}, n_{SMUTE})$
\STATE \textbf{return} $\mathcal{X}_{maj}'$, $\mathcal{X}_{min}'$
\end{algorithmic}
\end{algorithm}

\section{Experimental Study}

To evaluate the usefulness of the proposed approach we conducted an experimental study, with the goal of answering the following questions:
\begin{itemize}
    \item Can applying SMUTE alone lead to an improved classification performance when compared to other undersampling techniques?
    \item Can combining over- and undersampling in the form of CSMOUTE further improve the performance?
    \item Under what conditions is the performance improvement possible?
\end{itemize}
To answer these questions we performed an experimental study consisting of four stages, described in the remainder of this section.

\subsection{Set-up}

\noindent\textbf{Datasets.} Conducted experimental study was based on the binary imbalanced datasets provided in the KEEL repository \cite{alcala2011keel}. Their details are presented in Table~\ref{table:datasets}. In addition to the imbalance ratio (IR), the number of samples and the number of features, for each dataset we computed the proportion of different types of minority class observations, proposed by Napierała and Stefanowski \cite{napierala2016types}, which will be later used to analyse the impact of the dataset characteristics on the performance of the proposed method. Prior to resampling and classification, categorical features were encoded as integers. Afterwards, all features were standardized by removing the mean and scaling to unit variance. No further preprocessing was applied.

\begin{table*}
\small
\caption{Details of the datasets used during the experimental analysis.}
\label{table:datasets}
\centering
\setlength{\tabcolsep}{5pt}
\begin{tabular}{llllllll}
\toprule
\textbf{Name} & \textbf{IR} & \textbf{Samples} & \textbf{Features} & \textbf{Safe [\%]} & \textbf{Borderline [\%]} & \textbf{Rare [\%]} & \textbf{Outlier [\%]} \\
\midrule
glass1 & 1.82 & 214 & 9 & 47.37 & 28.95 & 17.11 & 6.58 \\
pima & 1.87 & 768 & 8 & 28.36 & 46.27 & 16.79 & 8.58 \\
glass0 & 2.06 & 214 & 9 & 54.29 & 38.57 & 2.86 & 4.29 \\
yeast1 & 2.46 & 1484 & 8 & 21.91 & 45.45 & 20.75 & 11.89 \\
haberman & 2.78 & 306 & 3 & 4.94 & 46.91 & 33.33 & 14.81 \\
vehicle1 & 2.9 & 846 & 18 & 23.04 & 57.6 & 14.75 & 4.61 \\
vehicle3 & 2.99 & 846 & 18 & 15.57 & 52.36 & 26.42 & 5.66 \\
ecoli1 & 3.36 & 336 & 7 & 53.25 & 31.17 & 7.79 & 7.79 \\
yeast3 & 8.1 & 1484 & 8 & 55.21 & 26.99 & 7.36 & 10.43 \\
ecoli3 & 8.6 & 336 & 7 & 28.57 & 48.57 & 8.57 & 14.29 \\
yeast-2\_vs\_4 & 9.08 & 514 & 8 & 54.9 & 19.61 & 7.84 & 17.65 \\
ecoli-0-6-7\_vs\_3-5 & 9.09 & 222 & 7 & 40.91 & 31.82 & 13.64 & 13.64 \\
yeast-0-3-5-9\_vs\_7-8 & 9.12 & 506 & 8 & 18.0 & 28.0 & 20.0 & 34.0 \\
yeast-0-2-5-6\_vs\_3-7-8-9 & 9.14 & 1004 & 8 & 34.34 & 30.3 & 14.14 & 21.21 \\
ecoli-0-2-6-7\_vs\_3-5 & 9.18 & 224 & 7 & 36.36 & 36.36 & 9.09 & 18.18 \\
yeast-0-5-6-7-9\_vs\_4 & 9.35 & 528 & 8 & 7.84 & 41.18 & 19.61 & 31.37 \\
ecoli-0-6-7\_vs\_5 & 10.0 & 220 & 6 & 40.0 & 40.0 & 5.0 & 15.0 \\
glass-0-1-6\_vs\_2 & 10.29 & 192 & 9 & 0.0 & 29.41 & 41.18 & 29.41 \\
ecoli-0-1-4-7\_vs\_2-3-5-6 & 10.59 & 336 & 7 & 65.52 & 17.24 & 0.0 & 17.24 \\
glass-0-1-4-6\_vs\_2 & 11.06 & 205 & 9 & 0.0 & 23.53 & 35.29 & 41.18 \\
glass2 & 11.59 & 214 & 9 & 0.0 & 23.53 & 47.06 & 29.41 \\
cleveland-0\_vs\_4 & 12.31 & 173 & 13 & 0.0 & 69.23 & 23.08 & 7.69 \\
yeast-1\_vs\_7 & 14.3 & 459 & 7 & 6.67 & 36.67 & 26.67 & 30.0 \\
glass4 & 15.46 & 214 & 9 & 30.77 & 46.15 & 0.0 & 23.08 \\
page-blocks-1-3\_vs\_4 & 15.86 & 472 & 10 & 64.29 & 25.0 & 7.14 & 3.57 \\
abalone9-18 & 16.4 & 731 & 8 & 4.76 & 23.81 & 16.67 & 54.76 \\
yeast-1-4-5-8\_vs\_7 & 22.1 & 693 & 8 & 0.0 & 6.67 & 43.33 & 50.0 \\
yeast-2\_vs\_8 & 23.1 & 482 & 8 & 55.0 & 0.0 & 10.0 & 35.0 \\
flare-F & 23.79 & 1066 & 11 & 4.65 & 37.21 & 32.56 & 25.58 \\
car-good & 24.04 & 1728 & 6 & 0.0 & 97.1 & 2.9 & 0.0 \\
car-vgood & 25.58 & 1728 & 6 & 32.31 & 67.69 & 0.0 & 0.0 \\
yeast4 & 28.1 & 1484 & 8 & 5.88 & 35.29 & 19.61 & 39.22 \\
winequality-red-4 & 29.17 & 1599 & 11 & 0.0 & 9.43 & 18.87 & 71.7 \\
yeast-1-2-8-9\_vs\_7 & 30.57 & 947 & 8 & 3.33 & 20.0 & 26.67 & 50.0 \\
yeast5 & 32.73 & 1484 & 8 & 31.82 & 54.55 & 9.09 & 4.55 \\
winequality-red-8\_vs\_6 & 35.44 & 656 & 11 & 0.0 & 0.0 & 44.44 & 55.56 \\
abalone-17\_vs\_7-8-9-10 & 39.31 & 2338 & 8 & 1.72 & 17.24 & 34.48 & 46.55 \\
abalone-21\_vs\_8 & 40.5 & 581 & 8 & 14.29 & 35.71 & 21.43 & 28.57 \\
yeast6 & 41.4 & 1484 & 8 & 37.14 & 25.71 & 11.43 & 25.71 \\
winequality-white-3\_vs\_7 & 44.0 & 900 & 11 & 0.0 & 15.0 & 5.0 & 80.0 \\
abalone-19\_vs\_10-11-12-13 & 49.69 & 1622 & 8 & 0.0 & 0.0 & 21.88 & 78.12 \\
kr-vs-k-zero\_vs\_eight & 53.07 & 1460 & 6 & 62.96 & 25.93 & 7.41 & 3.7 \\
winequality-white-3-9\_vs\_5 & 58.28 & 1482 & 11 & 0.0 & 8.0 & 12.0 & 80.0 \\
poker-8-9\_vs\_6 & 58.4 & 1485 & 10 & 4.0 & 60.0 & 20.0 & 16.0 \\
abalone-20\_vs\_8-9-10 & 72.69 & 1916 & 8 & 0.0 & 15.38 & 19.23 & 65.38 \\
kddcup-buffer\_overflow\_vs\_back & 73.43 & 2233 & 41 & 86.67 & 6.67 & 0.0 & 6.67 \\
poker-8-9\_vs\_5 & 82.0 & 2075 & 10 & 0.0 & 0.0 & 16.0 & 84.0 \\
poker-8\_vs\_6 & 85.88 & 1477 & 10 & 5.88 & 35.29 & 35.29 & 23.53 \\
kddcup-rootkit-imap\_vs\_back & 100.14 & 2225 & 41 & 68.18 & 22.73 & 0.0 & 9.09 \\
abalone19 & 129.44 & 4174 & 8 & 0.0 & 0.0 & 15.62 & 84.38 \\
\bottomrule
\end{tabular}
\end{table*}

\noindent\textbf{Classification.} Three different classification algorithms, representing different learning paradigms, were used throughout the experimental study. Specifically, we used multi-layer perceptron (MLP), support vector machine with RBF kernel (SVM) and logistic regression (LR). The implementations of the classification algorithms provided in the scikit-learn machine learning library \cite{pedregosa2011scikit} were used, and their default parameters remained unchanged.

\noindent\textbf{Evaluation.} For every dataset we reported the results averaged over the $5\times2$ cross-validation folds \cite{alpaydin1999combined}. To measure the classification performance in the imbalanced data setting we used F-measure, AUC and G-mean metrics. It is worth noting that F-measure tends to be more biased towards the majority class than the other two metrics, which was discussed by Brzezinski et al. \cite{brzezinski2019dynamics}.

\subsection{Results}

\noindent\textbf{Comparison of SMUTE and random undersampling.} We began our experiments with an examination of whether the proposed SMUTE algorithm achieves a competitive performance when compared to the baseline undersampling strategies. This step was designed as a safety check to ensure that any of the potential improvement in performance observed for CSMOUTE was not caused solely by combining SMOTE with any form of undersampling. To validate this hypothesis we compared the results of SMUTE and random undersampling (RUS) on all of the considered benchmark datasets, and used the Wilcoxon signed-rank test to examine the statistical significance of the observed differences. For both algorithms we performed the undersampling up to the point of achieving balanced class distribution. For SMUTE we used the number of nearest neighbors $k = 5$, fixed across the datasets. We present the results of the conducted experiment in Table~\ref{table:results-smute-rus}, in which we included both the number of datasets on which SMUTE achieved better, equal or worse performance, as well as the computed $p$-values. As can be seen, the results varied depending on the specific choice of classification algorithm and performance metric, with the latter having the biggest impact on the statistical significance of the results. While for all of the classifier and metric combinations using SMUTE led to achieving a better performance on a higher number of datasets than RUS, only with respect to the F-measure the results were statistically significant at the significance level $\alpha = 0.05$. Nevertheless, this is sufficient for the conclusion that using the proposed SMUTE algorithm can alone produce better results than RUS, in particular when the F-measure is used as the performance metric.

\begin{table}
\small
\caption{Results of a comparison between SMUTE and RUS. The number of wins and loses indicates the number of datasets on which SMUTE achieved, respectively, better and worse results than RUS.}
\label{table:results-smute-rus}
\centering
\begin{tabular}{llllll}
\toprule
 & Metric & Wins & Loses & Ties & $p$-value \\
\midrule
\multirow{3}{*}{SVM} & AUC & 30 & 20 & 0 & 0.3039 \\
 & F-measure & 36 & 14 & 0 & 0.0003 \\
 & G-mean & 28 & 22 & 0 & 0.4486 \\
\midrule
\multirow{3}{*}{MLP} & AUC & 27 & 23 & 0 & 0.9040 \\
 & F-measure & 34 & 16 & 0 & 0.0004 \\
 & G-mean & 28 & 22 & 0 & 0.7030 \\
\midrule
\multirow{3}{*}{LR} & AUC & 28 & 22 & 0 & 0.6675 \\
 & F-measure & 30 & 20 & 0 & 0.0375 \\
 & G-mean & 27 & 23 & 0 & 0.8811 \\
\bottomrule
\end{tabular}
\end{table}

\begin{figure}
\centering
\includegraphics[width=0.8\linewidth]{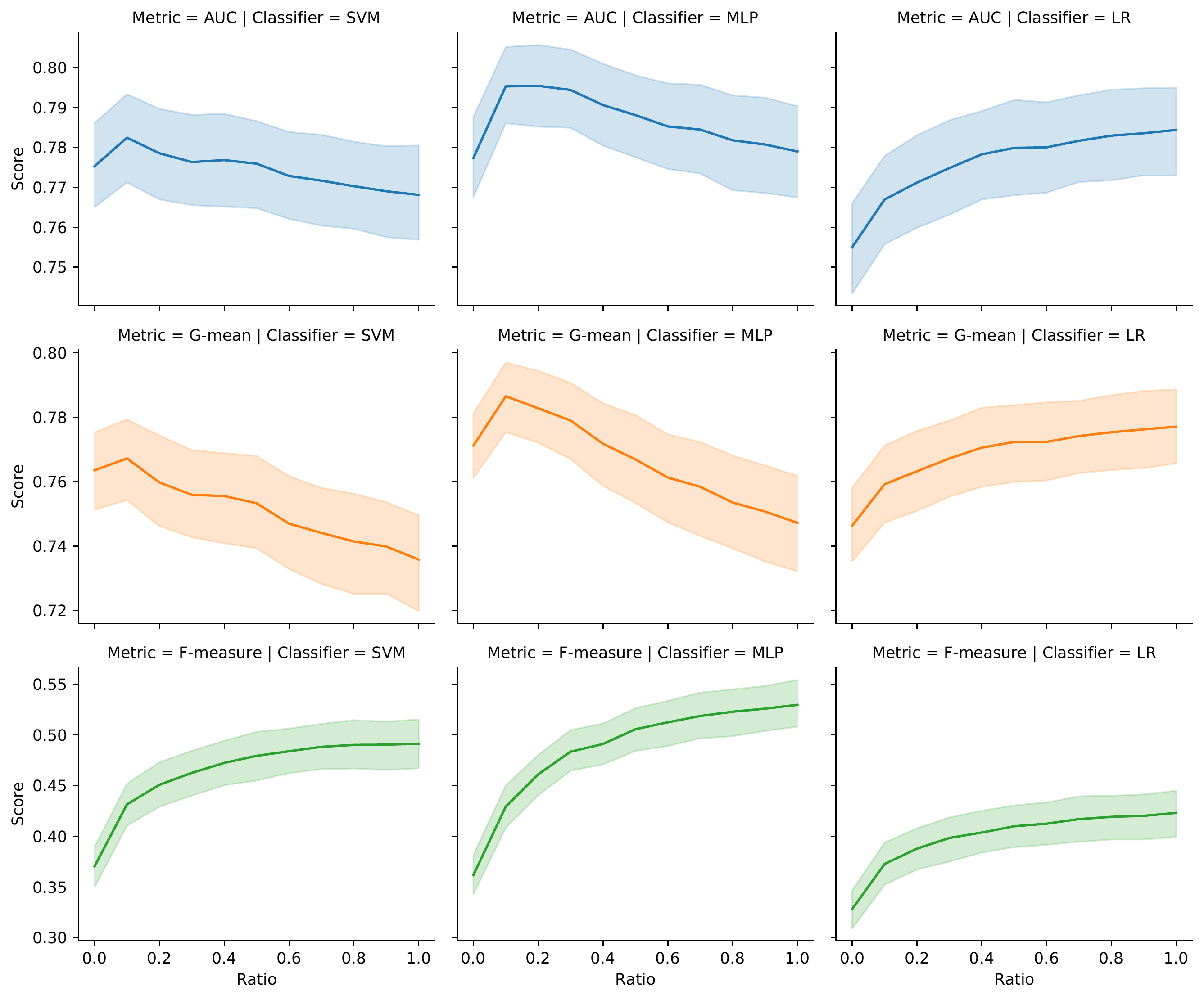}
\caption{The impact of ratio parameter on the CSMOUTE performance. The value of ratio equal to 1 corresponds to using SMOTE exclusively, whereas the value equal to 0 corresponds to using SMUTE exclusively. 95\% confidence intervals where shown.}
\label{fig:cmoute-ratio}
\end{figure}

\noindent\textbf{Analysis of the impact of oversampling ratio.} In the second stage of the conducted experimental study we examined the impact of the oversampling ratio of the CSMOUTE algorithm on its performance. To reiterate, the ratio corresponds to the proportion of data imbalance that is eliminated by using either over- or undersampling, with the ratio equal to 1 indicating that only SMOTE is used, and the ratio equal to 0 that only the SMUTE is used. In the conducted experiment we considered the values of oversampling ratio in $\in \{0.0, 0.1, 0.2, ..., 1.0\}$ and resampled up to the point of achieving balanced class distributions. For both SMOTE and SMUTE we used the number of nearest neighbors $k = 5$, fixed across the datasets. The results, averaged across the datasets, were presented in Figure~\ref{fig:cmoute-ratio}. As can be seen, once again the observed trends varied depending on the choice of the classification algorithm and the performance metric. However, two distinct types of response can be distinguished. First of all, the cases in which the underlying classification algorithm is LR, the performance metric is F-measure, or both. In those cases, on average, the observed performance decreased monotonically with the decrease of ratio. In other words, the optimal average performance was observed when SMOTE oversampling was used exclusively, and applying any undersampling negatively impacted the performance. Secondly, perhaps of a greater significance, the cases in which the classifier was either SVM or MLP, and the performance metric was either AUC or G-mean. In those cases a clear peak in average performance can be observed for small values of ratio parameter, or in other words when the majority of data rebalancing is being done using undersampling, but the oversampling is also used. The average performance improvement by combining over- and undersampling is largest in the case of MLP, which additionally achieved overall the best average performance out of the considered classification algorithms. Overall, the results indicate that it is possible to achieve an improvement in the performance by combining SMOTE with SMUTE, but on average this is the case only for a specific metrics and classifiers.

\begin{figure}
\centering
\includegraphics[width=\linewidth]{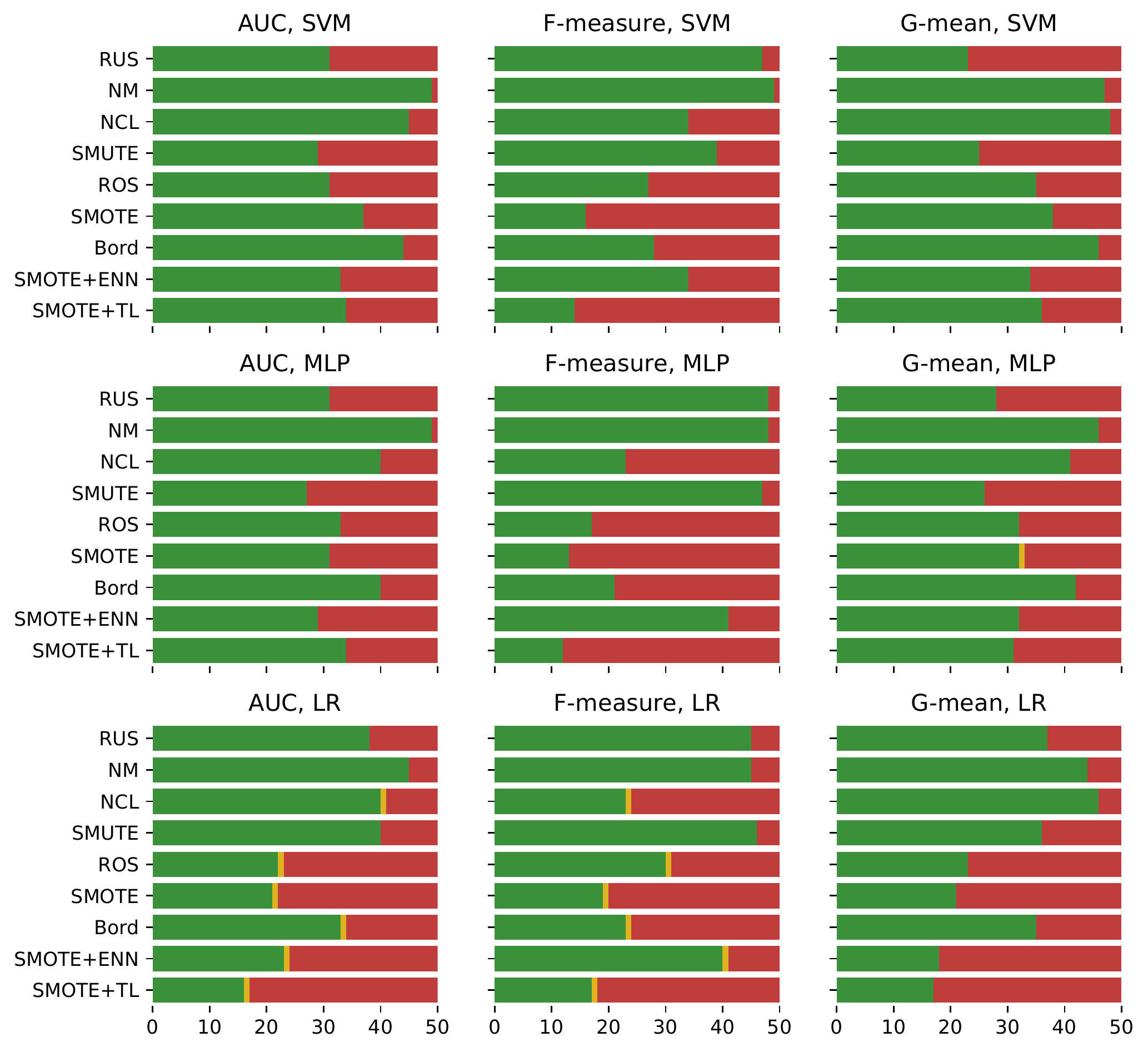}
\caption{A number of datasets on which CSMOUTE achieved better (green), equal (yellow) or worse (red) performance than individual reference methods.}
\label{fig:win-loss-tie}
\end{figure}

\noindent\textbf{Comparison of CSMOUTE and reference methods.} In the third stage of the conducted experimental study we compared the proposed CSMOUTE algorithm with a collection of a reference resampling strategies. We considered different over- and undersampling techniques, as well as the algorithms combining both of these approaches. Specifically, we used random undersampling (RUS), Near Miss method (NM) \cite{mani2003knn}, Neighborhood Cleaning Rule (NCL) \cite{laurikkala2001improving}, random oversampling (ROS), SMOTE \cite{chawla2002smote}, Borderline-SMOTE (Bord) \cite{han2005borderline}, and SMOTE combined with Tomek links (SMOTE+TL) \cite{tomek1976two} and Edited Nearest Neighbor rule (SMOTE+ENN) \cite{wilson1972asymptotic}. Furthermore, for reference we also included the proposed SMUTE algorithm. Whenever applicable, we conducted a parameter search using 2-fold cross-validation to select resamplers parameters individually for each dataset and performance metric. Specifically, for all variants of SMOTE, neighborhood-based reference undersampling strategies, that is NM and NCL, and during oversampling stage of CSMOUTE, we considered the values of $k$ neighborhood in \{1, 3, 5, 7\}. For Bord, we additionally considered the values of $m$ neighborhood, used to determine if a minority sample is in danger, in \{5, 10, 15\}. Finally, for both SMUTE and CSMOUTE we considered the values of $k$ neighborhood used for undersampling in \{1, 3, 5\}, and for the latter we considered the value of ratio parameter in \{0.0, 0.2, ..., 1.0\}. In all of the cases in which it was algorithms parameter the resampling was conducted up to the point of achieving balanced class distributions. To evaluate the statistical significance of the observed results we used the Friedman test combined with the Shaffer's post-hoc. The results were reported at the significance level $\alpha = 0.10$. Average ranks of the individual algorithms, with denoted cases in which given method achieved statistically significantly different results than CSMOUTE, were presented in Table~\ref{table:results-final}. Furthermore, a visualization of a win-loss-tie analysis, in which the number of datasets on which individual methods achieved better, worse or equal performance compared to CSMOUTE, was presented in Figure~\ref{fig:win-loss-tie}. As can be seen, CSMOUTE achieved a favorable performance with respect to the reference methods. The best results were observed for the scenarios in which combining SMOTE and SMUTE was previously shown to increase the overall performance, that is in combination with either SVM or MLP classifier, and using either AUC or G-mean as the performance metric. In those cases CSMOUTE achieved the highest average rank and, in specific cases, a statistically significantly better results than both the individual over- and undersampling reference strategies, including SMOTE and Bord. On the other hand, when the F-measure was used as the performance metric, statistically significantly better results were achieved only compared to the reference undersampling strategies. It must be noted that despite achieving a higher average rank in individual settings, in no case did CSMOUTE achieved a statistically significantly better performance than the reference strategies combining over- and undersampling, that is SMOTE+TL and SMOTE+ENN. On the other hand, it is also worth noting that the proposed approach did not achieve a statistically significantly worse results when compared to any of the other methods.

\begin{table*}
\scriptsize
\caption{Average ranks of individual methods. Reference algorithms achieving a statistically significantly worse results than CSMOUTE were denoted with a + sign (no algorithm achieved a statistically significantly better performance than CSMOUTE).}
\label{table:results-final}
\centering
\begin{tabularx}{\textwidth}{llYYYYYYYYYY}
\toprule
 & & \multicolumn{4}{l}{\textbf{Undersampling}} & \multicolumn{3}{l}{\textbf{Oversampling}} & \multicolumn{3}{l}{\textbf{Combined}} \\
\midrule
 & Metric & RUS & NM & NCL & SMUTE & ROS & SMOTE & Bord & SMOTE+ENN & SMOTE+TL & CSMOUTE \\
\midrule
\multirow{3}{*}{SVM} & AUC & 4.38 & 8.56 \textsubscript{+} & 8.20 \textsubscript{+} & 4.02 & 5.01 & 4.94 & 7.63 \textsubscript{+} & 4.24 & 4.68 & \textbf{3.34} \\
 & F-measure & 7.52 \textsubscript{+} & 9.18 \textsubscript{+} & 6.31 \textsubscript{+} & 6.62 \textsubscript{+} & 4.57 & 3.34 & 4.74 & 5.40 & \textbf{3.08} & 4.24 \\
 & G-mean & 3.46 & 8.02 \textsubscript{+} & 8.53 \textsubscript{+} & 3.72 & 5.37 \textsubscript{+} & 5.14 \textsubscript{+} & 7.92 \textsubscript{+} & 4.50 & 4.98 & \textbf{3.36} \\
\midrule
\multirow{3}{*}{MLP} & AUC & 4.98 & 8.92 \textsubscript{+} & 7.63 \textsubscript{+} & 4.20 & 4.62 & 4.82 & 7.19 \textsubscript{+} & 4.03 & 4.89 & \textbf{3.72} \\
 & F-measure & 8.10 \textsubscript{+} & 9.58 \textsubscript{+} & 4.61 & 7.64 \textsubscript{+} & 3.76 & \textbf{3.29} & 4.17 & 5.89 & 3.36 & 4.60 \\
 & G-mean & 4.34 & 8.30 \textsubscript{+} & 7.87 \textsubscript{+} & 4.32 & 4.76 & 5.06 & 7.35 \textsubscript{+} & 4.01 & 5.20 & \textbf{3.79} \\
\midrule
\multirow{3}{*}{LR} & AUC & 6.10 \textsubscript{+} & 8.46 \textsubscript{+} & 7.88 \textsubscript{+} & 6.68 \textsubscript{+} & \textbf{3.76} & 4.00 & 6.04 \textsubscript{+} & 3.94 & \textbf{3.76} & 4.38 \\
 & F-measure & 7.50 \textsubscript{+} & 8.82 \textsubscript{+} & 5.01 & 7.48 \textsubscript{+} & 4.70 & 3.90 & 3.99 & 5.74 & \textbf{3.68} & 4.18 \\
 & G-mean & 6.08 & 8.32 \textsubscript{+} & 8.47 \textsubscript{+} & 5.90 & \textbf{3.61} & 4.07 & 6.45 \textsubscript{+} & 3.81 & 3.83 & 4.46 \\
\bottomrule
\end{tabularx}
\end{table*}

\begin{figure}
\centering
\includegraphics[width=\linewidth]{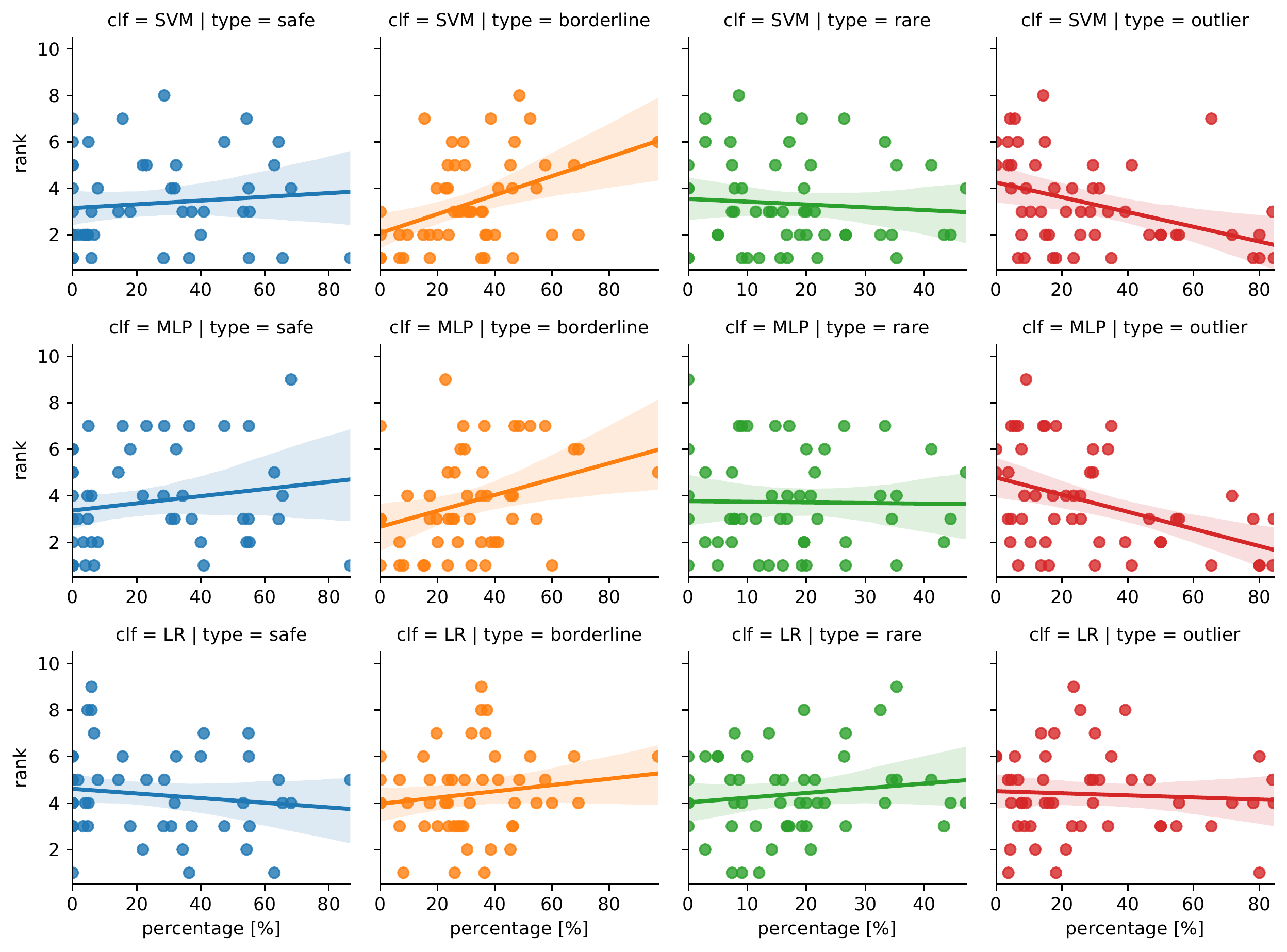}
\caption{A relation between the percentage of minority observations of different types in the given dataset and the rank achieved on it by CSMOUTE with respect to AUC metric. Four different types of observations were considered: safe (blue), borderline (orange), rare (green) and outlier (red). 95\% confidence intervals were shown.}
\label{fig:type-impact}
\end{figure}

\noindent\textbf{Analysis of the impact of dataset characteristics.} Finally, in the last stage of the conducted experimental analysis we considered the impact of the dataset characteristics on the overall performance of the proposed approach. We used a methodology proposed by Koziarski \cite{koziarski2020radial} to measure the correlation between the proportion of minority observations of a given type on the rank achieved on a specific dataset by the proposed CSMOUTE method. We used a classification proposed by Napierała and Stefanowski \cite{napierala2016types} which divides the observations into four types: safe, borderline, rare and outliers. The goal of this analysis was twofold: first of all, to identify the areas of applicability of the proposed approach, and give the user a heuristic that can be useful during determining if the method is likely to produce satisfactory results. Secondly, to facilitate a better understanding of the algorithms behavior and hopefully, outlining a future research direction. We present the results of our analysis in the form of Pearson correlation coefficients, with the statistically significantly correlations at the significance level $\alpha = 0.05$, in Table~\ref{table:results-type-impact}. Furthermore, we present scatterplots with linear regression fit to the data points with respect to the AUC metric in Figure~\ref{fig:type-impact}. As can be seen, a statistically significant correlation was observed precisely in the cases in which combining the over- and undersampling in form of CSMOUTE was shown to produce an improvement in the performance, that is for SVM and MLP classifiers and AUC and G-mean performance metrics. In those cases a trend can be observed in which the comparable performance of the proposed method increases with a high number of outlier instances, whereas it decreases with the number of borderline instances. This leads to a conclusion that the algorithm is particularly well suited for the datasets with a high number of outliers, such as datasets with a high level of label noise, at the same time displaying worse performance when applied on the borderline instances.

\begin{table}
\small
\caption{Pearson correlation coefficients between the proportion of minority objects of a given type in a dataset and the rank achieved on it by the CSMOUTE algorithm. Statistically significant correlations denoted with bold font. Note that negative correlation indicates increasing performance.}
\label{table:results-type-impact}
\centering
\begin{tabular}{llllll}
\toprule
 & Metric & Safe & Borderline & Rare & Outlier \\
\midrule
\multirow{3}{*}{SVM} & AUC & +0.102 & \textbf{+0.427} & -0.079 & \textbf{-0.408} \\
 & F-measure & -0.228 & +0.088 & +0.195 & +0.054 \\
 & G-mean & -0.085 & +0.271 & +0.018 & -0.145 \\
\midrule
\multirow{3}{*}{MLP} & AUC & +0.175 & \textbf{+0.321} & -0.015 & \textbf{-0.426} \\
 & F-measure & +0.228 & +0.096 & -0.122 & -0.240 \\
 & G-mean & +0.242 & +0.264 & -0.227 & \textbf{-0.338} \\
\midrule
\multirow{3}{*}{LR} & AUC & -0.135 & +0.150 & +0.141 & -0.061 \\
 & F-measure & +0.031 & +0.073 & +0.130 & -0.156 \\
 & G-mean & -0.274 & +0.100 & +0.133 & +0.120 \\
\bottomrule
\end{tabular}
\end{table}

\section{Conclusions}

In this paper we proposed a novel data-level algorithm for imbalanced data classification, SMUTE, an approach employing data interpolation present in SMOTE to reduce the number of majority observations. Furthermore, we extended it in form of CSMOUTE, a combined over- and undersampling strategy based on both SMOTE and SMUTE. We discussed the properties of CSMOUTE that make it suitable for further extension based on local data characteristics. Afterwards, we performed an experimental study in which we have shown that SMUTE is a viable alternative to a traditional undersampling strategies, and its performance can be further improved by combining with SMOTE in form of CSMOUTE. Finally, during the experimental analysis we were able to show that the performance of the algorithm shows a significant correlation with the characteristics of a dataset on which it is applied, with the performance improved on datasets consisting of a high proportion of outlier instances, and worsened on datasets with a high proportion of borderline instances. The fact that the algorithm displays a significantly different performance based on the global dataset characteristics suggest that it might be feasible to apply resampling selectively, based on the local characteristics of the data, to achieve a better performance. Specifically, a promising strategy might incorporate placing a particular focus on the outlier instances during the undersampling, at the same time treating borderline instances differently, for example by either excluding them completely or by applying oversampling instead. Additionally, the results of the conducted experiments show that the algorithms performance differs based on the underlying classification algorithm. Out of the considered classifiers, a statistically significant improvement in the performance was possible for MLP and SVM classifiers, with a greater improvement for the former, while the performance of LR was affected to a lesser extent. This might lead to a hypothesis that the performance of the proposed CSMOUTE approach is better when combined with a classifier producing a more complex decision boundaries, but further experimental analysis would be required to confirm it.

\section*{Acknowledgments}

This work was supported by the Polish National Science Center under the grant no. 2017/27/N/ST6/01705 as well as the PLGrid Infrastructure.

\bibliographystyle{IEEEtran}
\bibliography{bibliography}

\begin{thebibliography}{10}
\providecommand{\url}[1]{#1}
\csname url@samestyle\endcsname
\providecommand{\newblock}{\relax}
\providecommand{\bibinfo}[2]{#2}
\providecommand{\BIBentrySTDinterwordspacing}{\spaceskip=0pt\relax}
\providecommand{\BIBentryALTinterwordstretchfactor}{4}
\providecommand{\BIBentryALTinterwordspacing}{\spaceskip=\fontdimen2\font plus
\BIBentryALTinterwordstretchfactor\fontdimen3\font minus
  \fontdimen4\font\relax}
\providecommand{\BIBforeignlanguage}[2]{{%
\expandafter\ifx\csname l@#1\endcsname\relax
\typeout{** WARNING: IEEEtran.bst: No hyphenation pattern has been}%
\typeout{** loaded for the language `#1'. Using the pattern for}%
\typeout{** the default language instead.}%
\else
\language=\csname l@#1\endcsname
\fi
#2}}
\providecommand{\BIBdecl}{\relax}
\BIBdecl

\bibitem{koziarski2018convolutional}
M.~Koziarski, B.~Kwolek, and B.~Cyganek, ``Convolutional neural network-based
  classification of histopathological images affected by data imbalance,'' in
  \emph{Video Analytics. Face and Facial Expression Recognition}.\hskip 1em
  plus 0.5em minus 0.4em\relax Springer, 2018, pp. 1--11.

\bibitem{wei2013effective}
W.~Wei, J.~Li, L.~Cao, Y.~Ou, and J.~Chen, ``Effective detection of
  sophisticated online banking fraud on extremely imbalanced data,''
  \emph{World Wide Web}, vol.~16, no.~4, pp. 449--475, 2013.

\bibitem{azaria2014behavioral}
A.~Azaria, A.~Richardson, S.~Kraus, and V.~Subrahmanian, ``Behavioral analysis
  of insider threat: A survey and bootstrapped prediction in imbalanced data,''
  \emph{IEEE Transactions on Computational Social Systems}, vol.~1, no.~2, pp.
  135--155, 2014.

\bibitem{czarnecki2015compounds}
W.~M. Czarnecki and K.~Rataj, ``Compounds activity prediction in large
  imbalanced datasets with substructural relations fingerprint and {EEM},'' in
  \emph{2015 IEEE Trustcom/BigDataSE/ISPA}, vol.~2.\hskip 1em plus 0.5em minus
  0.4em\relax IEEE, 2015, pp. 192--192.

\bibitem{chawla2002smote}
N.~V. Chawla, K.~W. Bowyer, L.~O. Hall, and W.~P. Kegelmeyer, ``{SMOTE}:
  synthetic minority over-sampling technique,'' \emph{Journal of Artificial
  Intelligence Research}, vol.~16, pp. 321--357, 2002.

\bibitem{tomek1976two}
I.~Tomek, ``Two modifications of {CNN},'' \emph{IEEE Transactions on Systems,
  Man, and Cybernetics}, vol.~6, pp. 769--772, 1976.

\bibitem{wilson1972asymptotic}
D.~L. Wilson, ``Asymptotic properties of nearest neighbor rules using edited
  data,'' \emph{IEEE Transactions on Systems, Man, and Cybernetics}, vol.~2,
  no.~3, pp. 408--421, 1972.

\bibitem{hart1968condensed}
P.~Hart, ``The condensed nearest neighbor rule,'' \emph{IEEE transactions on
  information theory}, vol.~14, no.~3, pp. 515--516, 1968.

\bibitem{mani2003knn}
I.~Mani and I.~Zhang, ``k{NN} approach to unbalanced data distributions: a case
  study involving information extraction,'' in \emph{Proceedings of workshop on
  learning from imbalanced datasets}, vol. 126, 2003.

\bibitem{anand2010approach}
A.~Anand, G.~Pugalenthi, G.~B. Fogel, and P.~Suganthan, ``An approach for
  classification of highly imbalanced data using weighting and undersampling,''
  \emph{Amino acids}, vol.~39, no.~5, pp. 1385--1391, 2010.

\bibitem{smith2014instance}
M.~R. Smith, T.~Martinez, and C.~Giraud-Carrier, ``An instance level analysis
  of data complexity,'' \emph{Machine learning}, vol.~95, no.~2, pp. 225--256,
  2014.

\bibitem{yen2009cluster}
S.-J. Yen and Y.-S. Lee, ``Cluster-based under-sampling approaches for
  imbalanced data distributions,'' \emph{Expert Systems with Applications},
  vol.~36, no.~3, pp. 5718--5727, 2009.

\bibitem{beckmann2015knn}
M.~Beckmann, N.~F. Ebecken, B.~S.~P. de~Lima \emph{et~al.}, ``A {KNN}
  undersampling approach for data balancing,'' \emph{Journal of Intelligent
  Learning Systems and Applications}, vol.~7, no.~04, p. 104, 2015.

\bibitem{liu2008exploratory}
X.-Y. Liu, J.~Wu, and Z.-H. Zhou, ``Exploratory undersampling for
  class-imbalance learning,'' \emph{IEEE Transactions on Systems, Man, and
  Cybernetics, Part B (Cybernetics)}, vol.~39, no.~2, pp. 539--550, 2008.

\bibitem{galar2013eusboost}
M.~Galar, A.~Fern{\'a}ndez, E.~Barrenechea, and F.~Herrera, ``{EUSB}oost:
  Enhancing ensembles for highly imbalanced data-sets by evolutionary
  undersampling,'' \emph{Pattern Recognition}, vol.~46, no.~12, pp. 3460--3471,
  2013.

\bibitem{lu2017adaptive}
W.~Lu, Z.~Li, and J.~Chu, ``Adaptive ensemble undersampling-boost: a novel
  learning framework for imbalanced data,'' \emph{Journal of Systems and
  Software}, vol. 132, pp. 272--282, 2017.

\bibitem{drummond2003c4}
C.~Drummond, R.~C. Holte \emph{et~al.}, ``{C4.5}, class imbalance, and cost
  sensitivity: why under-sampling beats over-sampling,'' in \emph{Workshop on
  learning from imbalanced datasets II}, vol.~11.\hskip 1em plus 0.5em minus
  0.4em\relax Citeseer, 2003, pp. 1--8.

\bibitem{van2009knowledge}
J.~Van~Hulse and T.~Khoshgoftaar, ``Knowledge discovery from imbalanced and
  noisy data,'' \emph{Data \& Knowledge Engineering}, vol.~68, no.~12, pp.
  1513--1542, 2009.

\bibitem{garcia2012effectiveness}
V.~Garc{\'\i}a, J.~S. S{\'a}nchez, and R.~A. Mollineda, ``On the effectiveness
  of preprocessing methods when dealing with different levels of class
  imbalance,'' \emph{Knowledge-Based Systems}, vol.~25, no.~1, pp. 13--21,
  2012.

\bibitem{songwattanasiri2010smoute}
P.~Songwattanasiri and K.~Sinapiromsaran, ``{SMOUTE}: Synthetics minority
  over-sampling and under-sampling techniques for class imbalanced problem,''
  in \emph{Proceedings of the Annual International Conference on Computer
  Science Education: Innovation and Technology, Special Track: Knowledge
  Discovery}, 2010, pp. 78--83.

\bibitem{bunkhumpornpat2015core}
C.~Bunkhumpornpat and K.~Sinapiromsaran, ``{CORE}: Core-based synthetic
  minority over-sampling and borderline majority under-sampling technique,''
  \emph{International journal of data mining and bioinformatics}, vol.~12,
  no.~1, pp. 44--58, 2015.

\bibitem{junsomboon2017combining}
N.~Junsomboon and T.~Phienthrakul, ``Combining over-sampling and under-sampling
  techniques for imbalance dataset,'' in \emph{Proceedings of the 9th
  International Conference on Machine Learning and Computing}, 2017, pp.
  243--247.

\bibitem{laurikkala2001improving}
J.~Laurikkala, ``Improving identification of difficult small classes by
  balancing class distribution,'' in \emph{Conference on Artificial
  Intelligence in Medicine in Europe}.\hskip 1em plus 0.5em minus 0.4em\relax
  Springer, 2001, pp. 63--66.

\bibitem{napierala2016types}
K.~Napierala and J.~Stefanowski, ``Types of minority class examples and their
  influence on learning classifiers from imbalanced data,'' \emph{Journal of
  Intelligent Information Systems}, vol.~46, no.~3, pp. 563--597, 2016.

\bibitem{koziarski2020radial}
M.~Koziarski, ``Radial-{B}ased {U}ndersampling for imbalanced data
  classification,'' \emph{Pattern Recognition}, vol. 102, p. 107262, 2020.

\bibitem{han2005borderline}
H.~Han, W.-Y. Wang, and B.-H. Mao, ``Borderline-{SMOTE}: a new over-sampling
  method in imbalanced data sets learning,'' in \emph{International Conference
  on Intelligent Computing}.\hskip 1em plus 0.5em minus 0.4em\relax Springer,
  2005, pp. 878--887.

\bibitem{Bunkhumpornpat:2009}
C.~Bunkhumpornpat, K.~Sinapiromsaran, and C.~Lursinsap, ``{Safe-Level-SMOTE}:
  safe-level-synthetic minority over-sampling technique for handling the class
  imbalanced problem,'' in \emph{Advances in Knowledge Discovery and Data
  Mining, 13th Pacific-Asia Conference 2009, Bangkok, Thailand, April 27-30,
  2009, Proceedings}, 2009, pp. 475--482.

\bibitem{Maciejewski:2011}
T.~Maciejewski and J.~Stefanowski, ``Local neighbourhood extension of {SMOTE}
  for mining imbalanced data,'' in \emph{Proceedings of the {IEEE} Symposium on
  Computational Intelligence and Data Mining 2011, part of the {IEEE} Symposium
  Series on Computational Intelligence 2011, April 11-15, 2011, Paris, France},
  2011, pp. 104--111.

\bibitem{bunkhumpornpat2011mute}
C.~Bunkhumpornpat, K.~Sinapiromsaran, and C.~Lursinsap, ``{MUTE}: Majority
  under-sampling technique,'' in \emph{2011 8th International Conference on
  Information, Communications \& Signal Processing}.\hskip 1em plus 0.5em minus
  0.4em\relax IEEE, 2011, pp. 1--4.

\bibitem{saez2016analyzing}
J.~A. S{\'a}ez, B.~Krawczyk, and M.~Wo{\'z}niak, ``Analyzing the oversampling
  of different classes and types of examples in multi-class imbalanced
  datasets,'' \emph{Pattern Recognition}, vol.~57, pp. 164--178, 2016.

\bibitem{fernandez2018smote}
A.~Fern{\'a}ndez, S.~Garcia, F.~Herrera, and N.~V. Chawla, ``{SMOTE} for
  learning from imbalanced data: progress and challenges, marking the 15-year
  anniversary,'' \emph{Journal of artificial intelligence research}, vol.~61,
  pp. 863--905, 2018.

\bibitem{koziarski2019radial}
M.~Koziarski, B.~Krawczyk, and M.~Wo{\'z}niak, ``Radial-{B}ased {O}versampling
  for noisy imbalanced data classification,'' \emph{Neurocomputing}, vol. 343,
  pp. 19--33, 2019.

\bibitem{alcala2011keel}
J.~Alcal{\'a}-Fdez, A.~Fern{\'a}ndez, J.~Luengo, J.~Derrac, S.~Garc{\'\i}a,
  L.~S{\'a}nchez, and F.~Herrera, ``{KEEL} data-mining software tool: data set
  repository, integration of algorithms and experimental analysis framework.''
  \emph{Journal of Multiple-Valued Logic \& Soft Computing}, vol.~17, 2011.

\bibitem{pedregosa2011scikit}
F.~Pedregosa, G.~Varoquaux, A.~Gramfort, V.~Michel, B.~Thirion, O.~Grisel,
  M.~Blondel, P.~Prettenhofer, R.~Weiss, V.~Dubourg \emph{et~al.},
  ``Scikit-learn: Machine learning in {P}ython,'' \emph{Journal of Machine
  Learning Research}, vol.~12, no. Oct, pp. 2825--2830, 2011.

\bibitem{alpaydin1999combined}
E.~Alpaydin, ``Combined 5 $\times$ 2 cv {F} test for comparing supervised
  classification learning algorithms,'' \emph{Neural Computation}, vol.~11,
  no.~8, pp. 1885--1892, 1999.

\bibitem{brzezinski2019dynamics}
D.~Brzezinski, J.~Stefanowski, R.~Susmaga, and I.~Szcz{\k{e}}ch, ``On the
  dynamics of classification measures for imbalanced and streaming data,''
  \emph{IEEE transactions on neural networks and learning systems}, 2019.

\end{thebibliography}

\end{document}